\documentclass[conference]{IEEEtran}
\IEEEoverridecommandlockouts

\usepackage[numbers]{natbib}
\usepackage{amsmath,amssymb,amsfonts}
\usepackage{algorithmic}
\usepackage{graphicx}
\usepackage{textcomp}
\usepackage[table,dvipsnames]{xcolor}
\usepackage{todonotes}
\usepackage{hyperref}
\usepackage{diagbox}
\usepackage[T1]{fontenc}
\usepackage{pgfplots}
\usepackage{framed}

\definecolor{myred}{rgb}{1, 0.941, 0.941}
\definecolor{mygreen}{rgb}{0.941, 0.9725, 0.949}
\definecolor{myyellow}{rgb}{1, 0.988, 0.941}

\def\mybar#1{
  {\rule{#1in}{1ex}}}
\def\mysubbar#1{
 {\textcolor{Sepia!60}{\rule{#1in}{1ex}}}}

\def\BibTeX{{\rm B\kern-.05em{\sc i\kern-.025em b}\kern-.08em
    T\kern-.1667em\lower.7ex\hbox{E}\kern-.125emX}}

\begin{document}

\title{Systematic Mapping Study on the Machine Learning Lifecycle}

\author{\IEEEauthorblockN{Yuanhao Xie}
\IEEEauthorblockA{\textit{AI for Fintech Research, ING} \\
Amsterdam, Netherlands \\
yuanhao.xie@ing.com}
\and
\IEEEauthorblockN{Lu\'{i}s Cruz}
\IEEEauthorblockA{\textit{Delft University of Technology} \\
Delft, Netherlands \\
l.cruz@tudelft.nl}
\and
\IEEEauthorblockN{Petra Heck}
\IEEEauthorblockA{\textit{Fontys ICT} \\
Eindhoven, Netherlands \\
p.heck@fontys.nl}
\and
\IEEEauthorblockN{Jan S. Rellermeyer}
\IEEEauthorblockA{\textit{Delft University of Technology} \\
Delft, Netherlands \\
j.s.rellermeyer@tudelft.nl}
}

\maketitle

\begin{abstract}
The development of artificial intelligence (AI) has made various industries eager to explore the benefits of AI. There is an increasing amount of research surrounding AI, most of which is centred on the development of new AI algorithms and techniques. However, the advent of AI is bringing an increasing set of practical problems related to AI model lifecycle management that need to be investigated. We address this gap by conducting a systematic mapping study on the lifecycle of AI model. Through quantitative research, we provide an overview of the field, identify research opportunities, and provide suggestions for future research. Our study yields 405 publications published from 2005 to 2020, mapped in 5 different main research topics, and 31 sub-topics. We observe that only a minority of publications focus on data management and model production problems, and that more studies should address the AI lifecycle from a holistic perspective.
\end{abstract}


\section{Introduction}\label{sec:intro}

With the development of AI, various industries are eager to discover and reap the benefits of AI, and progressively more AI-related research is being carried out. However, the purpose of most of the research is to develop new AI algorithms or techniques to solve an issue in a given field. There are growing numbers of problems related to AI model life cycle management, such as version control of data and models, the difficulties of model deployment, transparency, model reproducibility, fairness, and so on~\cite{haakman2020ai,khomh2018software,arpteg2018software}. Despite this, there is a limited number of studies on AI model life cycle management, and there is no comprehensive study on the life cycle management of AI models. To fill this gap, we have conducted a systematic mapping study of the life cycle management of AI. By summarising the current situation in this field through quantitative and qualitative research we have developed an overview, identified research gaps and provided suggestions for future research directions.

 
We collect 405 papers and map them into 5 main categories with 31 sub categories. We also analyse the papers in terms of venue, year of publication, type of research, and 
Moreover, we discuss problems that are being overlooked by the AI research community.

The research questions are defined as:
RQ1: In what years, from which countries, affiliations, venue were these research papers published?
RQ2: What research approaches do these studies apply?
RQ3: Which subtopics of AI model life cycle management have already been investigated?

\subsection*{Replication Package}
All data (includes reference list of the 405 publications), instructions, necessary to replicate this study are available in the online appendix:\\
\url{https://luiscruz.github.io/ai-lifecycle-mapping-study/}


\section{Related Work}

In this section, we describe previous mapping studies that include the lifecycle of AI model applications in their scope.\\
Previous work conducted a systematic literature review to understand the use of software engineering practices in the development of AI/ML systems~\cite{nascimento2020software}. The study maps 57 ML studies into 11 different software engineering topics.~\cite{bourque2014guide}. It shows that little research has focused on deployment, maintenance, quality, and management. Our work differentiates by mapping the fields not only in terms of software engineering topics, but mostly by focusing on the different stages of the lifecycle of AI artefacts. 

\citet{vakkuri2018key} conceptualised and classified the ethics of AI, and found 37 keywords related to the ethics of AI in 83 papers. \citet{sherin2019systematic} conducted a system mapping study on the testing of machine learning systems. After researching 37 selected articles, they identified the trends in the field of machine learning testing and discovered research opportunities, such as testing machine learning programs using reinforcement learning. \citet{spaargaren2020systematic} conducted a systematic mapping study in the field of machine learning (ML) and software engineering (SE) in the context of financial technology (FinTech). None of the AI engineering related systematic mapping studies above are about AI model life cycle management, and only fragmented work about a specific sub topic of AI model lifecycle management. 

Other systematic mapping studies in the field of AI focus on the intersection between AI and other specific fields -- for instance, network security. \citet{wiafe2020artificial}.

\section{Methodology}\label{sec:methodology}
 
\begin{figure}
    \centering
    \includegraphics[width=0.8\linewidth]{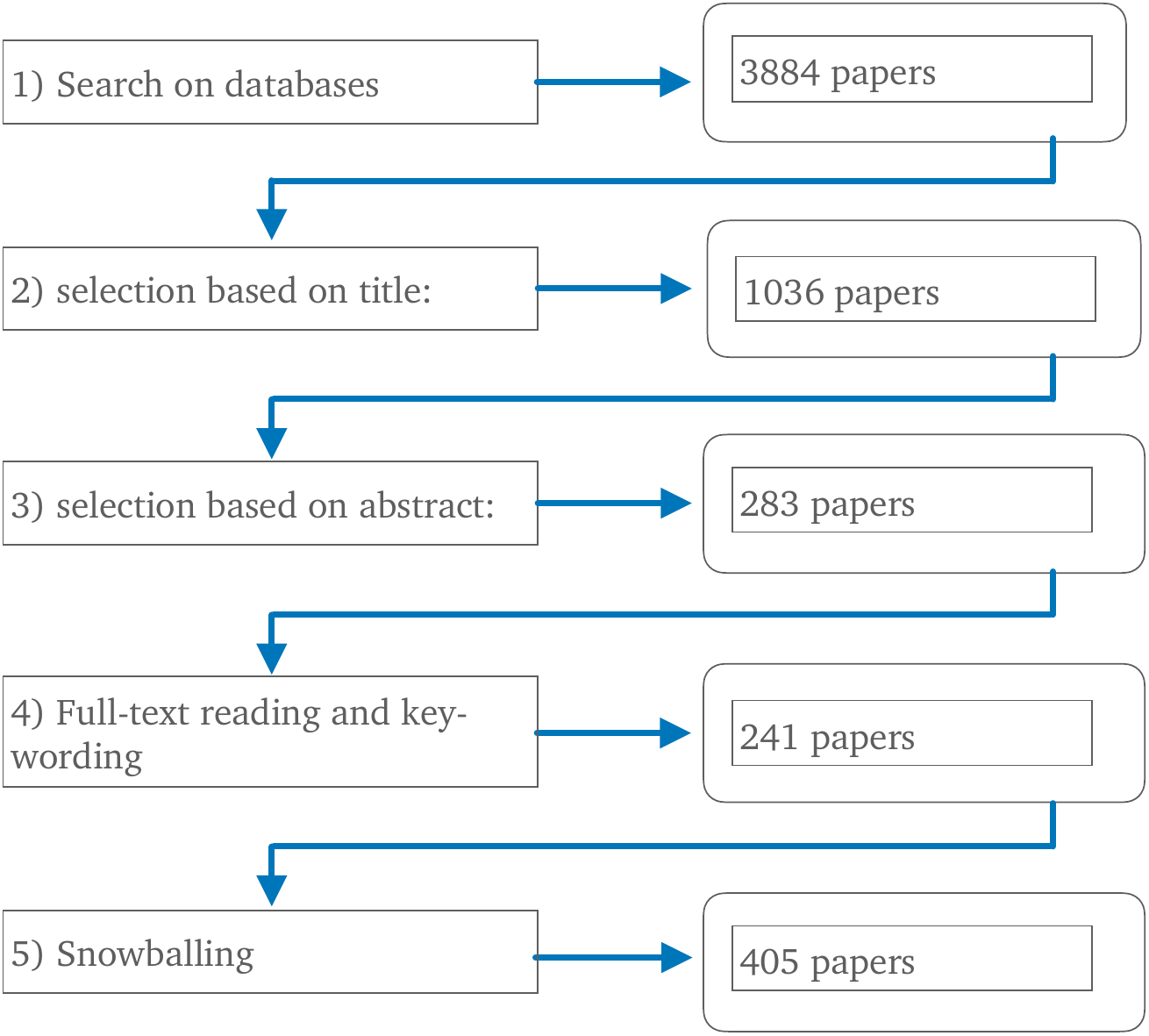}
    \caption{Methodology used to collect publications.}
    \label{fig:Steps of Methodology}
\end{figure}

 Our methodology, as described in Fig.~\ref{fig:Steps of Methodology}, is based on the guidelines proposed by Petersen et al. to conduct systematic mapping studies in software engineering~\cite{petersen2015guidelines,petersen2008systematic}. We pinpoint each step below.

\subsubsection{Search on Databases}
The search query we use to retrieve publications from the databases is illustrated in Fig. 2. We use the Population, Intervention, Comparison, and Outcome (PICO) strategy to derive the search terms. 

AI-related keywords includes \textbf{artificial intelligence (AI), machine learning (ML), deep learning (DL), neural network.}


We define keywords related to the lifecycle of AI based on previous work that pinpoints its different steps~\cite{rwortmanmorris,miao2017modelhub,kumeno2019sofware,datarobot}. A preliminary analysis yielded several keywords that cover three main aspects: model-oriented works (e.g., \textit{Pipeline}), code meets data (e.g., data \textit{Traceability}, \textit{Reproducibility}), and DevOps works (e.g., \textit{Deploy}). After analyzing the papers from the initial result set, we have iteratively improved the query by including, for example, \textit{Standard}, \textit{Best Practice}, and \textit{Platform}.

\begin{figure}
    \centering
    
    \includegraphics[width=\linewidth]{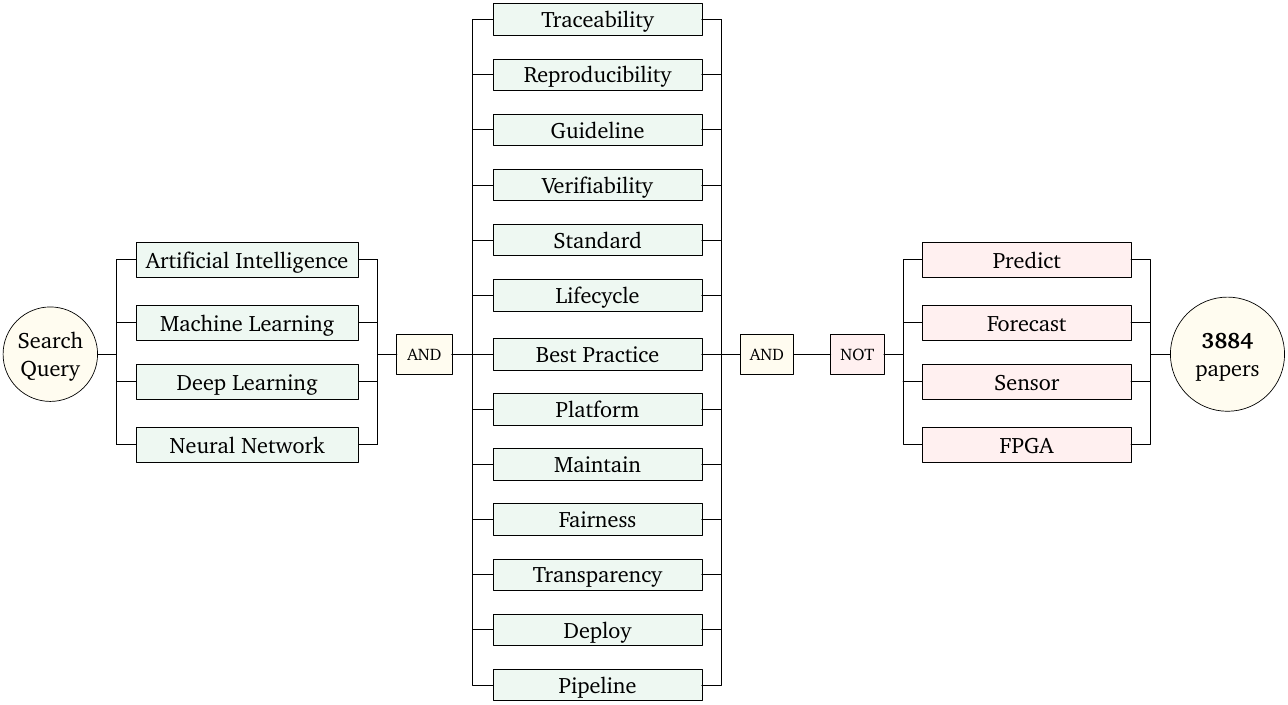}
    \caption{Diagram of the query used to retrieve papers from the DBLP and Scopus databases.}
    \label{fig:query}
\end{figure}



\subsubsection{Apply selection criteria based on title}\label{sec:select_title}

This step was undertaken by the first two authors in parallel. The papers were only excluded when both researchers agreed in that decision. In the end, we select 1036 papers,
The papers were excluded according to the following criteria:

\begin{itemize}
\item Studies not presented in English.
\item Duplicates.
\item Studies not accessible in full-text.
\item Books and grey literature.
\item Studies that discussed applications of AI.
\item Studies about the design of new AI algorithms unrelated to AI model lifecycle management.
\end{itemize}


\begin{table}[t]
    \centering
    \caption{Agreement table of the selection based on title.}
    \scalebox{0.9}{
    \begin{tabular}{l l  r  r  r}
                    &                                 & \multicolumn{3}{c}{Researcher 1}                                  \\
                    &                                 & Include                   & Uncertain                 & Exclude   \\
\hline
                    & Include                         & \cellcolor{mygreen}195   & \cellcolor{myyellow}67   & \cellcolor{myyellow}71        \\
    Researcher 2    & Uncertain                       & \cellcolor{myyellow}129  & \cellcolor{myyellow}85   & \cellcolor{myyellow}212       \\
                    & Exclude                         & \cellcolor{myyellow}119  & \cellcolor{myyellow}158  & \cellcolor{myred}\textbf{2848}      \\
    \end{tabular}}
    \label{tab:title_stage}
\end{table}

\subsubsection{Apply selection criteria based on abstract.}
We exclude all the papers in which one of the researchers proposed to exclude and the other agreed or was uncertain -- yielding 628 papers, as stated in the \textbf{red cells highlighted in bold}.



\begin{table}[]
    \centering
    \caption{Agreement table of the selection based on abstract.}
    \begin{tabular}{l l  r  r  r}
                    &                                 & \multicolumn{3}{c}{Researcher 1}                                  \\
                    &                                 & Include                   & Uncertain                 & Exclude   \\
\hline
                    & Include                         & \cellcolor{mygreen}253   & \cellcolor{mygreen}6   & \cellcolor{myyellow}38        \\
    Researcher 2    & Uncertain                       & \cellcolor{mygreen}31  & \cellcolor{myyellow}21   & \cellcolor{myred}\textbf{6}       \\
                    & Exclude                         & \cellcolor{myyellow}57  & \cellcolor{myred}\textbf{12}  & \cellcolor{myred}\textbf{610}      \\
    \end{tabular}
    \label{tab:abs_stage}
\end{table}

\subsubsection{Full-text reading and Keywording.}

In this step, the first author goes through the contents of the 283 papers to remove articles that were out of scope. As a result,  241 papers are remaining.



\subsubsection{Snowballing}
The first step is to browse the reference list, read the abstracts of the articles in the list, exclude papers that do not meet the basic criteria (such as language, publication year, and publication type), and then delete duplicate articles. After this step, 164 articles were obtained.
\section{Results}
\label{sec:results}
In this section, we present all the results collected in this mapping study. We map 405 papers according to year, countries, affiliations, venue and lifecycle topic. We also provide a fine-grained mapping in each subtopic of the lifecycle.
\subsection{RQ1: In what years, from which countries, affiliations, venue were these research papers published?}
\subsubsection{Publication by country}
The country is determined according to the institution where first author is located. The top five countries are the United States(49.4\%), Germany(8.4\%), the United Kingdom(6.2\%), China(5.9\%), and Canada(4.2\%), while other countries accounted for less than 3\%.
Publications from the United States account for about half of the total. This trend is not uncommon~\cite{barricelli2019end}, but it emphasises the lack of diversity and the importance of having more countries shaping the future of AI.

\subsubsection{Publication by year}
The number of studies on the lifecycle management of AI models has increased over time. Especially from 2016 to 2019, the number of publications each year is double that of the previous year. The publications in this mapping study were collected in the first half of 2020, so the 2020 publication information is not complete.


\subsubsection{Publication by affiliations}
270 different companies, universities or organizations have published research on the life cycle management of artificial intelligence models. Eight of the universities/companies produced more than five publications. IBM has produced 22 articles,
half of them are about risk management. Google produced 11 articles, five main themes are all included. University of Maryland, Stanford University, Carnegie Mellon University, University of California, and University of Oxford produced less than 10 related publications.


\subsubsection{Publication by type of venue}
65.4\% of all publications are conference papers, 17.5\% are journal articles, 14.6\% are informative and other publications, and the remaining 10 publications are parts in books or collections.

\subsection{RQ2: What research approaches do these studies apply?}
\subsubsection{Publication by type of research type}
We map papers according to its research type using the taxonomy proposed in~\cite{wieringa2006requirements}. It includes six different types: Solution, Evaluation, Philosophical, Validation, Opinion, Experience. The distribution is presented in Table~\ref{tab:research_type}.


\begin{table}[]
    \centering
    \caption{Distribution of papers across research types.}
    \scalebox{1.0}{
    \begin{tabular}{l r r l}
    Research Type   & Count     & Percentage (\%)   & \\
    \hline
    Solution        &      186     & 45.9                &\mybar{.459}            \\
    Evaluation       &      74     & 18.3              &\mybar{.183}           \\
    Philosophical   &    73       & 18.0                &\mybar{.180}             \\
    Validation      &     56      & 13.8                &\mybar{.138}             \\
    Opinion         &     14      &  3.5                &\mybar{.035}             \\
    Experience      &      2     &  0.5                &\mybar{.005}            
    \end{tabular}}
    \label{tab:research_type}
\end{table}

Solution papers were by far the most common research type, comprising 45.9\% of the publications.
This means that most works focus on proposing a novel solution technique with a proof of concept consisting of a small example or a sound argument.
Evaluation research follows with 18.3\%, comprising papers that investigate a problem or the implementation of a technique in practice.
Not so significant yet representative are Philosophical papers with 18.0\% and Validation papers with 13.8\%.

\subsection{RQ3: Which subtopics of AI model life cycle management have already been investigated?
}

\subsubsection{Publication by research topic}

We first summarised the keywords of each article, and after reading all the articles, the keyword database is obtained. Then we read all the articles again, and labelled each article by selecting the suitable keywords from the keyword database. All the articles are grouped into 5 overarching research topics and 31 themes (sub-topics) as presented in Table~\ref{tab:category}.
Amongst which, "Risk Management" works that address governance and management of risks associated with AI. This is a broad topic, but here we exclusively focus on works that study risk management from the perspective of the lifecycle of the AI system. "Lifecycle management" is a separate category with no subcategories, containing all articles discussing artificial intelligence management from a holistic perspective.

    
   



\begin{table}[]
    \centering
    \caption{Distribution of papers amongst topic.}
    \scalebox{1.0}{
    \begin{tabular}{l r r l}
    \textsc{Topic}/Subtopic         & Count     & \%   & \\
    \hline
    \rowcolor{gray!10}
    \textsc{- Risk management}          & 143       &35.3               &\mybar{.353}            \\
    Security                        & 61        &15.1               &\mysubbar{0.151}           \\
    Fairness                        & 44        &10.9               &\mysubbar{0.109}             \\
    Transparency                    &  14       &3.5                &\mysubbar{0.035}     \\
    Privacy                         &  9        &2.2                &\mysubbar{0.022}     \\
    Reproducibility                 &  7        &1.7                &\mysubbar{0.017}     \\
    Ethics                          &  6        &1.5                &\mysubbar{0.015}     \\
    Risk management                     &  2        &0.5                &\mysubbar{0.005}     \\

    \rowcolor{gray!10}
    \textsc{- Model Management}     & 138       &34.1               &\mybar{0.341}           \\
    Explainability                  &  49       &12.1                &\mysubbar{0.121}             \\
    Interpretation                  &  24       &5.9                &\mysubbar{0.059}     \\
    Visualization                   &  18       &4.4                &\mysubbar{0.044}     \\
    Development                     &  16       &4.0               &\mysubbar{0.040}     \\
    Evaluation                      &  9       &2.2                &\mysubbar{0.022}     \\
    AutoML                          &  4       &1.0                &\mysubbar{0.010}     \\
    Experiment Management           &  4       &1.0               &\mysubbar{0.010}     \\
    Model Management                &  4       &1.0                &\mysubbar{0.010}     \\
    Hyperparameter Optimization/Management                         
                                    &  3       &0.7         
    &\mysubbar{0.010}     \\
    Sharing                         &  2       &0.5 &           \mysubbar{0.005}     \\
    Model Traceability/Versioning   &  2       &0.5 &          
    \mysubbar{0.005}      \\
    Training Management             &  2       &0.5 &           \mysubbar{0.0050}     \\
    Technical Debt                  &  1       &0.2 &           \mysubbar{0.002}     \\

    \rowcolor{gray!10}
    \textsc{- Production}       & 59        &14.6               &\mybar{.146}             \\
    Deployment                     &  31       &7.7                &\mysubbar{0.077}     \\
    Testing                  & 22        &5.4              &\mysubbar{0.054}           \\
    Make use of AI                   &   6      &1.5                &\mysubbar{0.015}     \\

    \rowcolor{gray!10}
    \textsc{- Lifecycle Management}   &  46       &11.4               &\mybar{.114}             \\
    Lifecycle Management            &  46       &11.4                &\mysubbar{0.114}     \\

    \rowcolor{gray!10}
    \textsc{- Data Management}      &  19       &4.7                &\mybar{.047}     \\
    Data preprocessing/preparation  &  7        &1.7                &\mysubbar{0.017}     \\
    Data Cleaning                   &  3        &0.7                &\mysubbar{0.007}     \\
    Data Management                 &  3        &0.7                &\mysubbar{0.007}     \\
    Annotation                      &  2        &0.5                &\mysubbar{0.005}     \\
    Data Traceability/Versioning    &  2        &0.5                &\mysubbar{0.005}     \\
    Extraction, Transform, Load     &  1        &0.2                &\mysubbar{0.002}     \\
    Quality Assessment              &  1        &0.2                &\mysubbar{0.002}     \\

    \end{tabular}}
    \label{tab:category}
\end{table}

\section{Discussion}

The majority of papers collected in this study tackle a particular stage of the AI model lifecycle management. Albeit relevant, we argue that it is important look into the lifecycle of AI model from a holistic perspective -- only 46 papers (11.4\%) in our study took such perspective. On the other hand, this suggests that there are still many problems to be solved in each particular stage. Our study also shows that some sub-categories have been overlooked by the research community -- e.g., \textit{Data Traceability/Versioning}, \textit{Quality Asessment}.

In terms of research type, a considerable portion of the publications (45.9\%) revolve around proposing a \textit{Solution}. The nature of the problems related to the lifecycle management of AI projects is mostly related to how to manage AI projects in practice, in a real setting. Thus, we argue that more \textit{Evaluation} papers are necessary for this topic. Thus, strong collaborations between academia and the industry are key to improve the state of the art of the lifecycle management of AI applications.

\textit{Opinion} and \textit{Experience} publications were the least found research types -- 3.5\% and 0.5\%, respectively. This is expected, as academic venues tend to be strict with these types of works that are subject to author bias. However, there are no standards on the AI lifecycle and engineers are figuring out how to apply it in practice. Hence, we argue that \textit{Opinion} and \textit{Experience} papers are also very important to the community of AI engineers. This is a concern being raised by the research community of empirical software engineering, which highlights grey literature as a valuable data source for research~\cite{garousi2020benefitting}.

We also noticed that when referring to specific topics, many articles are confused about topics with similar meanings. For example there are usually two kinds of papers about explanation: using mathematical models to explain ML models, and explaining model features. In the process of reading the literature, we found that many documents confuse explainability and interpretability. This may be due to the lack of a clear and authoritative definition of those topics, because the current research on the life cycle of artificial intelligence is still in its infancy.

\section{Threats to Validity}

\subsection{Theoretical validity}






\subsubsection{Study Identification/Sampling, Data extraction and classification}

The search query is based on common keywords that may appear in the title or abstract. While we made extensive efforts to iteratively improve the search query, relevant papers may have been missed. This is a common threat in mapping studies~\cite{wohlin2013reliability}. To mitigate this threat, we complement the search with backward snowball sampling. Furthermore, to reduce the researcher bias, two authors performed the paper selection. The threat is not fully mitigated since only the first author classified the selected papers in the last step of selection.

\subsection{Generalisability}

Generalisability is not a major threat since this mapping study cover a wide range of topics. Yet, from 2009 to 2019, the number of papers is increasing almost exponentially every year. Thus, new publications in subsequent years may substantially affect the quantitative results of this mapping study. We argue that this is a common threat in any mapping study done in a fast-changing field, but does not hinder their usefulness.

\subsection{Interpretive validity}

The conclusions drawn from the data collected are prone to researcher bias. We mitigate this threat by having four researchers discussing and reviewing the interpretation of data.

\subsection{Reproducibility and Repeatability}

We report the systematic mapping process followed. Furthermore, all the scripts and data collected are delivered the online appendix of this mapping study (cf. Section~\ref{sec:intro}).

\section{Conclusion}

In this paper, we map 405 research publications published from 2005 to 2020 in 5 different main research topics and 31 sub topics related to the lifecycle management of AI systems. All the data and analysis is available online in the replication package (cf. Section~\ref{sec:intro}).


As future work, we plan to extend this research to publications in 2020 and 2021, and perform a systematic literature review.

\section*{Acknowledgement}

We would like to thank Jerry Brons and Elvan Kulan for their valuable feedback in this work. This study was supported by the ICAI lab \href{https://se.ewi.tudelft.nl/ai4fintech/index.html}{AI for Fintech Research}.
\footnotesize{
\bibliographystyle{IEEEtranN}
\bibliography{refs}
}
\end{document}